\documentclass{article}

\usepackage[numbers]{natbib}
\usepackage[preprint]{neurips_2020
}

\usepackage[utf8]{inputenc} 
\usepackage[T1]{fontenc}    
\usepackage{hyperref}       
\usepackage{url}            
\usepackage{booktabs}       
\usepackage{amsfonts}       
\usepackage{microtype}      
\usepackage{graphicx}
\usepackage{amsmath}
\usepackage{csquotes}

\title{Generative Language Modeling for Automated Theorem Proving}

\author{
  Stanislas Polu \\
  OpenAI \\
  \texttt{spolu@openai.com} \\
  \And
  Ilya Sutskever \\
  OpenAI \\
  \texttt{ilyasu@openai.com} \\
}

\date{August 2020}

\begin{document}

\maketitle

\begin{abstract}
 We explore the application of transformer-based language models to automated theorem proving.  This work is motivated by the possibility that a major limitation of automated theorem provers compared to humans -- the generation of original mathematical terms -- might be addressable via generation from language models. We present an automated prover and proof assistant, \emph{GPT-f}, for the Metamath formalization language, and analyze its performance. \emph{GPT-f} found new short proofs that were accepted into the main Metamath library, which is to our knowledge, the first time a deep learning based system has contributed proofs that were adopted by a formal mathematics community.

\end{abstract}

\section{Introduction}

Artificial neural networks have enjoyed a spectacularly successful decade, having made considerable advances in computer vision~\cite{krizhevsky2012imagenet,he2016deep}, translation~\cite{sutskever2014sequence,bahdanau2014neural,wu2016google}, speech recognition~\cite{graves2014towards,amodei2016deep}, image generation \cite{goodfellow2014generative,radford2015unsupervised,karras2017progressive,karras2019style,chen2020generative}, game playing~\cite{silver2016mastering,berner2019dota,vinyals2019grandmaster}, and robotics~\cite{akkaya2019solving,levine2016end}.  Especially notable is the recent rapid progress in language understanding and generation capabilities~\cite{vaswani2017attention,radford2018improving,radford2019language,brown2020language,devlin2018bert}. 

With the possible exception of AlphaGo~\cite{silver2016mastering} and AlphaZero~\cite{silver2017mastering}, reasoning tasks are conspicuously absent from the list above. In this work we take a step towards addressing this absence by applying a transformer language model to automated theorem proving.

Automated theorem proving \cite{harrison2014history} is an appealing domain for exploring reasoning in general and the reasoning capabilities of language models in particular for several reasons:

\begin{itemize}
    \item {\bf Reasoning-complete:} Proving theorems very likely require general and flexible reasoning;  thus an advance in theorem proving is also an advance in reasoning more broadly.
    \item {\bf Search:} Automated theorem proving systems can quickly check the correctness of proofs, making it a productive environment for the use and development of search methods. 
    \item {\bf Automated data generation:} The ability to verify proofs makes it possible to automatically generate new problems that could then be used as training data. This is especially important, since collecting high quality data for reasoning tasks can be difficult.
\end{itemize}

Learning to prove theorems is somewhat analogous to learning to play Go:  both offer an automated way of determining success (the game of Go is a miniature formal system), and both offer an automated way for generating new data via self play-type approaches.  This similarity, together with the clear success of AlphaZero, suggests that automated theorem proving might prove to be a fruitful domain for the study of reasoning in neural networks where significant progress may be possible.

\newpage
\subsection{Contribution}

Our contributions are the following:
\begin{itemize}

\item We verify that generative pre-training substantially improves performance and that pre-training on mathematical data (such as arXiv) leads to better performance compared to pre-training on generic text from the web. 
\item We find that model size is positively correlated with performance, even though the size  of the Metamath dataset is relatively small. 
\item We demonstrate that iteratively training a value function on statements generated by our language model leads to improved prover performance, which immediately suggests a strategy for continuous self improvement: keep training on proofs generated by the prover.
\item We also achieve a new state of the art for the Metamath environment with our best model capable of closing $56.22\%$ of proofs from a held-out \textit{test} set (vs $21.16\%$ for the current state of the art, \textit{MetaGen-IL}~\cite{wang2020learning}), demonstrating that the Transformer architecture may be suitable to formal reasoning.
\end{itemize}

\section{Related Work}

\paragraph{Deep learning applied to premise selection and proof guidance} Research on automated theorem proving dates back to the 50s~\cite{harrison2014history}, but mainstream proof assistants still suffer from combinatorial explosion of their search space as they are scaled to large corpuses, motivating the use of deep learning. Early applications of deep learning to formal mathematics focused primarily on premise selection and proof guidance. DeepMath~\cite{irving2016deepmath} explored the use of CNNs and RNNs to predict whether a premise is useful to demonstrate a given conjecture, their results were later improved with FormulaNet~\cite{wang2017premise} by the use of graph neural networks, reminiscent of NeuroSAT~\cite{selsam2018learning}. Proof guidance consists in selecting the next clause to process \textit{inside} an automated theorem prover. Loos et al.~\cite{loos2017deep} investigated the use of models similar to DeepMath's for proof guidance and demonstrated a significant uplift on the Mizar library.

\paragraph{Deep learning applied to automated theorem-proving} \textit{HOList}~\cite{bansal2019holist} proposes a formal environment based on HOL Light. They achieve their best performance~\cite{bansal2019learning} with a GNN model designed for premise selection and the use of exploration. More recently, the same team studied the use of the BERT objective with Transformers on formal statements~\cite{rabe2020language}, demonstrating the potential of leveraging Transformers for formal reasoning. Their study focuses on preliminary tasks that are related but not directly consisting of proving formal theorems (such as typing and conjecturing). \textit{GamePad}~\cite{huang2018gamepad} and \textit{CoqGymn/ASTactic}~\cite{yang2019learning} introduce environments based on the Coq theorem prover. \textit{ASTactic} generates tactics as programs by sequentially expanding a partial abstract syntax tree. \textit{Holophrasm}~\cite{whalen2016holophrasm} and \textit{MetaGen-IL}~\cite{wang2020learning} propose RNN-based models to generate proofs for Metamath (the formal system we focus on). They rely on three different models, one to value goals, one to select premises and one to generate substitutions. \textit{MetaGen-IL} also demonstrates an uplift in performance by generating synthetic data by forward proving. 

\paragraph{Use of Transformers for symbolic tasks} Several lines of work have been exploring language modeling using Transformers~\cite{vaswani2017attention}. Language modeling improvements have been demonstrated from better pre-training tasks, using various objectives such as auto-regressive generation~\cite{radford2018improving,radford2019language,brown2020language}, token masking~\cite{devlin2018bert} or sequence masking~\cite{raffel2019exploring}, but resulting language models have so far felt short when applied to reasoning oriented tasks such as algebraic word problems~\cite{ling2017program,amini2019mathqa}. Recently, Lample and Charton~\cite{lample2019deep} successfully applied Transformers to anti-derivative calculus and solving differential equations, hinting that Transformers are capable of generating the exogenous terms involved in the substitutions required for successful symbolic integration. The Universal Transformer~\cite{dehghani2018universal}, a Transformer with tied weights, was also shown to be successful at more algorithmic tasks. Also, Saxton et al.~\cite{saxton2019analysing} evaluated the Transformer architecture on a variety of mathematical problems.

\section{Formal Environment}

We chose Metamath~\cite{megill2019metamath} as our formal environment. Metamath is powered by a simple meta logic system based on a single substitution rule~\cite{megill2006how}.

The main Metamath library is called \textit{set.mm}, which is a collection of $\sim 38k$ proofs based on ZFC set theory (while other formalisms can also be used on top of Metamath's meta logic, they are not used in {\it set.mm}).

Metamath has several advantages that make it convenient to use with neural networks:
\begin{itemize}
  \item Verification is fast and can be implemented in several hundreds lines of code.
  \item Proof steps are \textit{context-free}: a goal or subgoal that we wish our system to prove, together with a list of the statements of the theorems proven so far, completely define the state of the Metamath system at any stage of a proof. Other formal systems are generally wrapped in high-level programming languages that make them easier to use for humans (by including convenient features like module imports or custom user-defined tactics) but are harder to integrate with a neural network. While proofs in such systems are generally shorter and more human-readable, they are impacted by long-distance interactions which makes the complete description of the intermediary states of proofs longer, and therefore less suitable for neural language models.
  \item Access to clean and compact subgoal representations makes searching the proof tree relatively straightforward. It is not the case for systems where the proving objective resembles program synthesis more than an explicit proof tree search.
  \item \textit{set.mm} is one of the largest libraries available and its foundations are accepted as compatible with modern mathematics.
\end{itemize}

But it also has a number of weaknesses:
\begin{itemize}
  \item Metamath does not have high-level tactics, which means that all of its proof steps are very low-level. As an example, the \textit{de-bruijn factor}~\cite{wiedijk2014debruijn}--the quotient of the size of a formalization of a mathematical text and the size of its informal original version-- of a Metamath proof is $\sim10-20$ while it is around $\sim1-3$ in Coq, HOL Light or Lean.  Lower level proof steps mean longer proofs with greater chance of compounding errors during search.
  \item The current state of the tooling around Metamath makes it a very ``\textit{DIY}'' system, one that is not yet ready for broad adoption by the mathematics community.
\end{itemize}

While our approach would be applicable to other formal systems (such as Lean, Coq, or HOL Light), Metamath's features allow faster prototyping and reduced iteration time in the near term, which is why we chose it for this project.

The \textit{set.mm} library contains the background theorems required to demonstrate most Olympiad or undergraduate Mathematics type of problems. For example, assisted by the \emph{GPT-f} proof assistant described in this work in section~\ref{section:gptfproofassistant}, we formalized IMO 1972 problem B2\footnote{\textit{Metamath Proof Explorer - imo72b2} \url{http://us.metamath.org/mpeuni/imo72b2.html}}.

\subsection{Proving in Metamath}

Proving in Metamath consists of applying a previously demonstrated theorem or axiom by providing a substitution of the variables appearing in the hypotheses and conclusion of the theorem being applied, such that the substituted conclusion unifies to (which means that it "matches") the current goal which we wish to prove.  The substituted hypotheses, if any, become the new subgoals left to prove.

This mechanism, a \textit{proof step}, can be used in a forward manner (where we start with the axioms and reach the desired statement, one proof step at a time) and a backward manner (where we start with the statement we wish to prove and, after applying enough proof steps, end up at axioms or previously demonstrated theorems with whose hypothesis we already determined to be true). As it is more naturally amenable to proof search, we will be operating backward. 

As an example, assume we want to prove $\vdash ( 3 + 2 ) = 5$ using the definition of $4$ and $5$ as respective successors of $3$ and $4$. As a first step, we should use an equality transitivity theorem such as:

\begin{verbatim}
[[
  |- A = B            # first hypothesis
  |- C = B            # second hypothesis
]]
|- A = C              # conclusion
\end{verbatim}

To apply the transitivity theorem, we need to provide a substitutions that substitutes $A$ with $( 3 + 2)$ and $B$ with $5$ such that the conclusion of the theorem unifies to the current goal. We are left with providing a substitution for $B$ which can hardly be discovered mechanically (hence the appeal to use generative language modeling). We can substitute $B$ with $( 4 + 1 )$ as is the case in the actual proof\footnote{\textit{Metamath Proof Explorer - 3p2e5} \url{http://us.metamath.org/mpeuni/3p2e5.html}} in Metamath's \textit{set.mm} library.

Putting it all together, the goal here is:

\begin{verbatim}
|- ( 3 + 2 ) = 5
\end{verbatim}

The proof step we apply:

\begin{verbatim}
[[
  |- A = B            # first hypothesis
  |- C = B            # second hypothesis
]]
|- A = C              # conclusion
{{ A : ( 3 + 2 ) }}   # substitution of A
{{ B : ( 4 + 1 ) }}   # substitution of B
{{ C : 5 }}           # substitution of C
\end{verbatim}

And finally the new subgoals are:

\begin{verbatim}
|- ( 3 + 2 ) = ( 4 + 1 )
|- ( 4 + 1 ) = 5
\end{verbatim}

Applying the following proof step with no hypothesis (the definition of $5$\footnote{\textit{Metamath Proof Explorer - df-5} \url{http://us.metamath.org/mpeuni/df-5.html}}) to the second subgoal allows us to prove it.

\begin{verbatim}
[[ ]] |- ( 4 + 1 ) = 5
\end{verbatim}

Note that this proof step has no hypothesis and no substitution involved. It therefore closes that branch of the proof tree. From there the proof can be continued with the first subgoal, proving backward, until no subgoal is left. Also note that a proof for a given theorem of the library can only use theorems proven before the appearance of the theorem to prove; we enforce that constraint when benchmarking our models despite them being trained on the library as a whole.

In most formal systems, a \textit{proof step}, consists of a \textit{goal} and a mechanism that, given a goal produces new subgoals, generally referred to as a \textit{tactic}. In Metamath, there is only one type of \textit{tactic} based on substitution as illustrated above. Additionally since the substituted theorem must unify to the current goal, the current \textit{goal} can be deduced from the \textit{tactic} itself (theorem and substitution pair), which is not generally the case in other systems. As such, we'll use \textit{tactic} and \textit{proof step} interchangeably in the rest of the paper.

This informal presentation of Metamath is sufficient to understand the objectives we use to train our models. A more formal definition of Metamath's meta-logic can be found in the Metamath Book~\cite{megill2019metamath}.

\subsection{Dataset}

Metamath's \textit{set.mm} uses a binary compressed format to represent proofs of statements. We process the library and extract a dataset of proof steps, stored as JSON blobs using the representation presented above. For each proof step we store a \textit{GOAL}, a \textit{PROOFSTEP} and a reference to the parent goal if any, encoding the tree structure of the proofs:

\begin{verbatim}
{
  "proof_label": "unidmrn",
  "goal": "[[ ]] |- U. U. `' A = ( dom A u. ran A )",
  "proof_step": "[[ |- A = B |- C = B ]] |- A = C \\
     {{ A : U. U. `' A }} \\
     {{ B : ( ran `' A u. dom `' A ) }} \\
     {{ C : ( dom A u. ran A ) }}",
  "proof_step_hash": "37yZVNorgF8=",
  "parent_hash": ["n4Kl7judEN4="]
}
\end{verbatim}

The dataset contains $\sim 3m$ of such proof steps for $\sim 38k$ theorems (different proof labels). We split that dataset between a \textit{train} set and two \textit{valid} and \textit{test} sets each containing $\sim 1k$ proofs sampled randomly ($\sim 90k$ proof steps each).

\subsection{Glossary}

\begin{tabular}{p{3cm} p{10.1cm}} 
  \textit{term} & A string that comply to the Metamath grammar. \\[4pt]
  \textit{statement} or \textit{proposition} & A potentially empty set of hypotheses (terms) and a conclusion (term) entailed by the hypotheses. \\[4pt]
  \textit{theorem} & A proven statement. \\[4pt]
  \textit{axiom} & An assumed statement. \\[4pt]
  \textit{goal} & A statement in the context of a proof search. \\[4pt]
  \textit{substitutions} & A list of pairs of variables and terms (to substitute the variables within a theorem or an axiom). \\[4pt]
  \textit{tactic} & A theorem and substitutions that unify to a goal. \\[4pt]
  \textit{subgoals} & Goals generated by a tactic (the substituted hypotheses of the tactic's theorem). \\[4pt]
  \textit{proof step} & A goal and a tactic, potentially generating new subgoals. \\[4pt]
  \textit{proof} & A tree of goals and tactics whose root is the demonstrated theorem; leaves of the tree are tactics with no subgoals or goals that are hypotheses of the root theorem. \\[4pt]
\end{tabular}

\section{Model}

\subsection{Architecture}

We use decoder-only Transformers similar to GPT-2~\cite{radford2019language} and GPT-3~\cite{brown2020language}. The largest model we study has 36 layers and 774m trainable parameters.

\subsection{Training Objective}

The \textit{proofstep objective} we use for training is a conditional language modeling objective that is asked to generate the \textit{PROOFSTEP} given a \textit{GOAL}, which is directly applicable to proof searches. To do so, we format our data in the following way:

\begin{verbatim}
GOAL <GOAL> PROOFSTEP <PROOFSTEP><EOT>
\end{verbatim}

There is one such objective for each JSON line in our dataset. We train with only one sentence per context (no-chunking), masking the rest of the context by assigning a loss weight $w_{\mathit{loss}}=0$. As we train we track the valid loss and sequence accuracy while masking the query part of the objective:

\begin{verbatim}
GOAL <GOAL> PROOFSTEP
\end{verbatim}

We regularize the training by early-stopping at the point of minimum valid loss and applying a weight decay $\mathit{wd}=0.1$.

Here is a randomly sampled context as presented to our models for training:

\begin{verbatim}
GOAL [[ ]] |- ( ( ( J e. Nrm /\ f e. ( J Homeo K ) ) /\ ( x e. K /\  y e. ( 
( Clsd ` K ) i^i ~P x ) ) ) -> ( `' f " x ) e. J ) PROOFSTEP [[ |- ( ph -> 
ps ) |- ( ph -> ch ) |- ( ( ps /\ ch ) -> th ) ]]  |- ( ph -> th ) {{ ch : 
x e. K }} {{ ph : ( ( J e. Nrm /\ f e.  ( J Homeo K ) ) /\ ( x e. K /\ y e. 
( ( Clsd ` K ) i^i ~P x ) ) ) }}  {{ ps : f e. ( J Cn K ) }} {{ th : ( `' f 
" x ) e. J }} <|endoftext|>
\end{verbatim}

\subsection{Proof Search}

\subsubsection{Goal Expansion}
\begin{figure}
    \includegraphics[width=\textwidth]{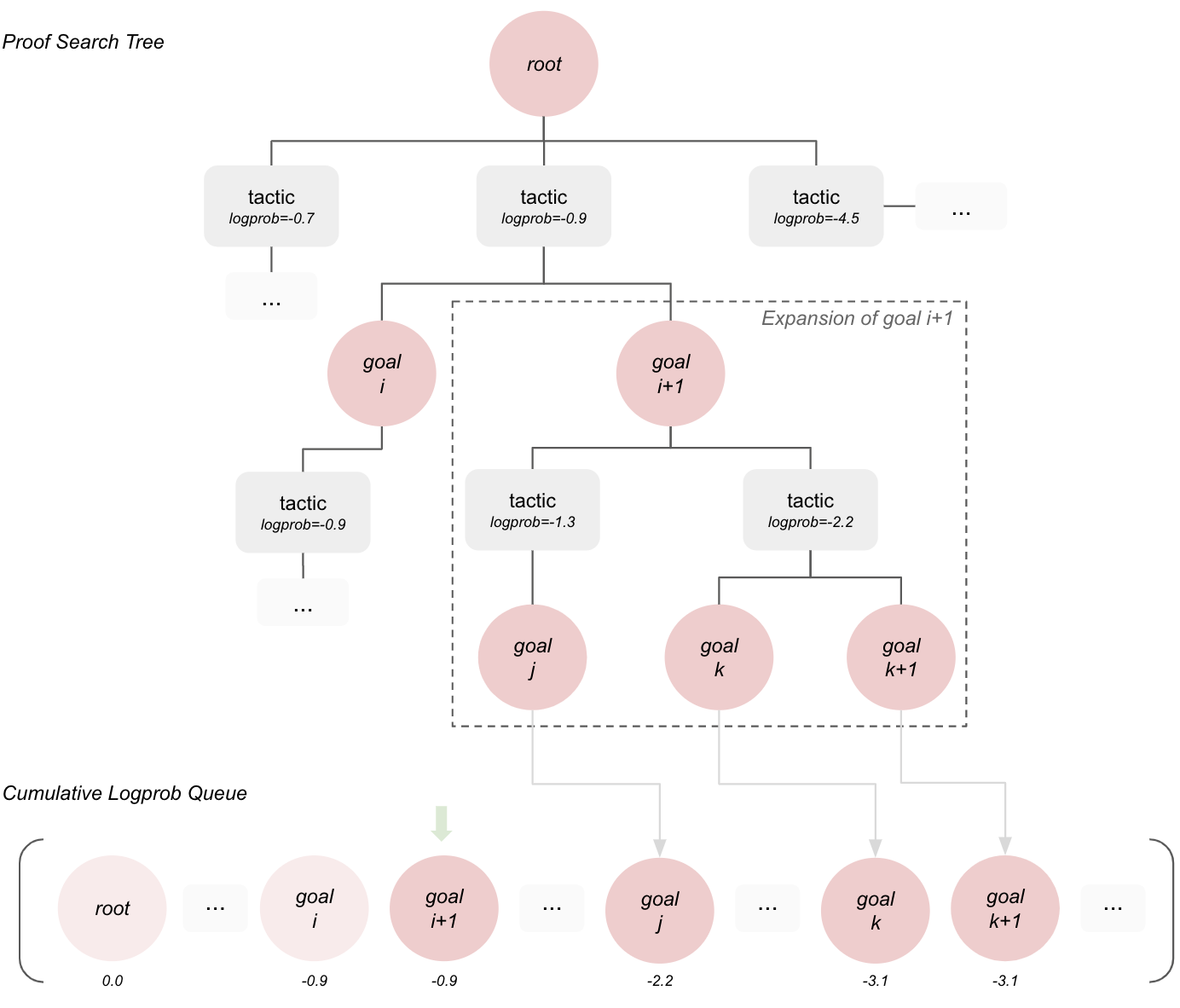}
    \caption{Proof search consists in maintaining a proof tree where multiple tactics are explored for each goal, starting from the root goal. Goals are expanded by cumulative (tactic) logprob priority.}
    \label{fig:proofsearch}
\end{figure}

We find proofs by running proof searches. A proof search maintains a proof tree and a queue of open goals sorted by their cumulative logprob, initialized with the root goal that we wish to demonstrate (see figure~\ref{fig:proofsearch}). The cumulative logprob of a goal is defined by the sum of the logprobs of the tactics that were used to reach that goal from the root goal. Intuitively we expand goals for which the generative model is the most confident globally. This has a tendency to explore breadth first as deeper goals have more parent tactics and therefore typically a higher cumulative logprob.

Each time we expand an open goal we sample $e=32$ tactics (the \textit{proofstep objective} described above) from the model at temperature $t=1.0$, deduplicate them, and apply the valid tactics (of which there are at most $e$) to the goal being expanded. Each successful tactic application generates new subgoals that are added to the proof tree and the proof search queue. The expanded goal is then removed from the queue. Note that the subgoals associated with a successfully applied tactic all share the same cumulative logprob and will eventually be expanded together (as subgoals generated from their own expansion will mechanically have a higher cumulative logprob, and will therefore be inserted behind in the queue). We denote the process of selecting the minimal cumulative logprob goal and expanding it as a proof search \textit{expansion}. 

Each proof search involves $d=128$ goal expansions, so proofs generated have at most $d$ proof steps. When evaluating our models, we attempt a proof search for each statement in the \textit{valid} set $a=4$ times, starting from an empty proof tree each time. In the above, $a$, $e$, and $d$ are hyperparameters of the search process that we can vary to achieve better performance (at the cost of more compute), but keep constant as we compare models.

\subsubsection{Formal Verifier}
Performing such proof searches requires to tightly couple a Metamath verifier with our models. We implemented a Metamath kernel in Python to avoid the performance cost and brittleness of interacting with an external kernel over its REPL through standard I/O. It also provides us with a flexible environment to experiment with new ideas in ways that were not anticipated by existing verifiers. The kernel consists of a modified \textit{LR(0)} parser to check that terms generated by our models comply with the Metamath grammar, along with \textit{Goal} and \textit{Tactic} objects that implement the Metamath substitution and represent proof trees. Our implementation is capable of exporting our in-memory representations to both our JSON marshalled format and the official \textit{set.mm} proof format. The latter allows us to verify the proofs we generate with an external Metamath kernel implementation such as \verb|mmverify.py| or \verb|metamath-exe|.

Collectively, this proof search procedure and the formal verifier tied with it are what we refer to as the \textit{GPT-f} automated prover.

\subsection{Evaluation}

We report the performance $\text{Perf}_{a,e,d}^{\mathit{valid}}(\theta)$ of a model $\theta$, as the percentage of proofs found by this procedure within the \textit{valid} or \textit{test} set. We evaluate our models on the \textit{valid} set of $\sim 1k$ theorems and once at the end of this paper on the held-out \textit{test} set.

The hyperparameters we chose to evaluate our models attempt to minimize the variance in the evaluation process while using the least amount of compute. Decreasing the number of expansions per goal $e$ increases the
variance as we less systematically explore the action space when expanding each goal. The $e$ value can be taken quite high as each auto-regressive proof step generation uses the same query for a given goal and can therefore be batched together. Increasing $e$ too much may also hurt performance given the breadth first nature of the cumulative logprob ordering. Increasing the number of attempts per proposition $a$ decreases variance and consistently improves performance up to reasonably high values (We use $a=32$ attempts for our final benchmarking). We found that $a=4$ limited the variance in our evaluations while remaining tractable given the amount of compute we had available. Finally the proof search depth $d$ has little impact on variance but naturally improves performance (we take $d=128$ to evaluate our models and $d=256$ for our final benchmarking).

The number of expansions we use per proof search may appear as relatively low, but it's important to realize that it already requires a substantial amount of compute as each expansion consists in the auto-regressive generation of $e=32$ tactics (generally hundreds of tokens and therefore forward passes each). Empirically, the hyperparameters we chose, require on average around $\sim 1k$ GPU.hours (with V100s) to evaluate our 700m parameters model (which leverages GPT-3's sparse attention as well as key-value caching).

\subsection{Pre-training}

We study the effect of pre-training on the performance of our models. We pre-train our models on both GPT-3's post-processed version of CommonCrawl as well as a more reasoning-focused mix of Github, arXiv and Math StackExchange.

Github is downloaded using BigQuery\footnote{\url{https://console.cloud.google.com/marketplace/details/github/github-repos}} and filtered to only include deduplicated files from selected programming languages (excluding markdown, stylesheets, HTML). arXiv is downloaded using Bulk Data Access\footnote{\url{https://arxiv.com/help/bulk\_data}} and filtered to only include articles labeled as Mathematics and whose LaTeX source is available. Math StackExchange is downloaded from their snapshot on the Internet Archive\footnote{\url{https://archive.org/details/stackexchange}} and post-processed to remove HTML tags and correlate questions and answers. We demote the mix reported in the table below as \textit{WebMath}:

\begin{table}[ht]
\caption{Mix and source of data involved in the \textit{WebMath} dataset.}
\centering
\begin{tabular}{ |l|c|c| }
    \hline
    Dataset & Size & Mix \\
    \hline
    Github & 23 GB & 33\% \\
    arXiv Math & 10 GB & 33\% \\
    Math StackExchange & 2 GB & 33\% \\
    \hline
\end{tabular}
\label{table:webmathmix}
\end{table}

\subsection{Synthetic Datasets}

Despite being among the largest formal mathematics libraries, the Metamath library remains scarce in the context of deep learning, especially in light of the advantages demonstrated on various NLP tasks by pre-training on large corpora. Also \textit{set.mm} mostly focuses on well-known high-level theorems and does not include a large number of technical lemmas resembling the type of mathematics exercises used as curriculum for humans. Finally, Metamath lacking high level tactics such as HOL Light's \verb|ARITH_RULE|\footnote{\url{https://www.cl.cam.ac.uk/~jrh13/hol-light/HTML/ARITH_RULE.html}}, or Lean's \verb|ring|\footnote{\url{https://leanprover-community.github.io/mathlib_docs/algebra/ring/basic.html\#ring}}, it is critical to ensure that our models are capable of proving at least basic technical theorems generally handled by high-level tactics in other systems (in domains such as arithmetic or ring equalities and inequalities)

To achieve this goal we designed synthetic datasets allowing us to generate proofs for each of these domains at will while controlling precisely by how many proofs we augment our training set.

We describe below the synthetic datasets we designed and report in section~\ref{section:experiments} the sample complexity associated with these synthetic tasks.

\subsubsection{\textit{n-digit} Arithmetic}

We synthetically generate proofs for arithmetic formulas such as $11*22=242$ by following the basic algorithm for addition and multiplication, repeatedly applying theorems such as \verb|decadd|\footnote{\textit{Metamath Proof Explorer - decadd} \url{http://us.metamath.org/mpeuni/decadd.html}} or \verb|decaddc|\footnote{\textit{Metamath Proof Explorer - decaddc} \url{http://us.metamath.org/mpeuni/decaddc.html}}. Divisions and subtractions are translated to their equivalent additions and multiplications theorems in one proof step. We also support generating modulos and exponentiations.

We accept one hyperparameter for these synthetic proof generators, $\mathit{ndigits}$ which controls the number of digits involved in these arithmetic tasks. When generating a new proof we sample uniformly in $[-10^{\mathit{ndigits}}, 10^{\mathit{ndigits}}]$ each of the numbers involved. To illustrate the level at which Metamath operates, Table~\ref{table:proofsteps} shows the average number of proof steps generated as a function of $\mathit{ndigits}$ for each generator. These statements are generally proved with one tactic application in other higher-level systems, which is a good example of one of Metamath's drawbacks we identified earlier.

\begin{table}[ht]
\caption{Average number of proofsteps produced by our synthetic generators for $\mathit{ndigits}=3,9,18$.}
\centering
\begin{tabular}{ |l|c|c|c| }
    \hline
     & 3 & 9 & 18 \\
    \hline
    Addition (in $\mathbb{Z}$) & 19 & 48 & 94 \\
    Division & 13 & 93 & 292 \\
    Modulo & 25 & 82 & 206 \\
    Exponentiation & 7 & 27 & 68 \\
    \hline
\end{tabular}
\label{table:proofsteps}
\end{table}

Our goal is to leverage these synthetic generators to ensure our models are confident when faced with such subgoals in order to mitigate the large number of proof steps they require.

\subsubsection{Ring Algebra}

Our ring equalities generator is largely inspired by the INT inequality generator~\cite{wu2020int} . They propose an inequality generator that starts from simple formulas (such as $A=A$) and iteratively transforms them into more complex equalities or inequalities using a predefined list of axioms (such as commutativity of addition or distributivity of addition-multiplication). At each transformation, the axiom to be applied is chosen uniformly.

Our generator operates similarly within the Metamath formalism based on theorems equivalent to the axioms they propose. We accept two hyperparameters, the number of variables $\mathit{nbvar}$ involved in the seed formulas (of the form $A=A$) as well as the number of theorems applied to transform the expression, denoted as $\mathit{depth}$. In addition, we use hand-crafted weights as we sample theorems in order to obtain formulas that we judged qualitatively better.

Here is a list of the theorems we use and their associated sampling weights.

\begin{table}[ht]
\caption{Metamath theorems use by our Ring Algebra synthetic generators. Theorems are available in the Matmath Proof Explorer.}
\centering
\begin{tabular}{ |l|c|l| }
    \hline
    Theorem & Weight & Description \\
    \hline
    \verb|eqcomd| & 1 & Commutative law for class equality. \\
    \verb|int-addcomd| & 1 & Addition commutativity. \\
    \verb|int-addassocd| & 1 & Addition associativity. \\
    \verb|int-mulcomd| & 1 & Multiplication commutativity. \\
    \verb|int-mulassocd| & 1 & Multiplication associativity. \\
    \verb|int-leftdistd| & 3 & Left distribution of multiplication over addition. \\
    \verb|int-rightdistd| & 3 & Right distribution of multiplication over addition. \\
    \verb|int-sqdefd| & 5 & Definition of the square. \\
    \verb|muladdd2| & 5 & Product of two sums \\
    \hline
\end{tabular}
\label{table:theorems}
\end{table}

Examples of equalities produced by the generator:

\begin{align*}
ABBA(AB)^2+(C+A) &= A+(ABBA)^2+C \\
(AA)^2 &= A^2AA \\
((BA+CA)^2)^2 &= (BA+CA)^2(BAAB+ACCA+BAAC+ABCA)\\
((A+B)^2)^2(A+A) &= ((A+B)^2(AB+AB+AA+BB)+ (A+B)^2(AB+AB+AA+BB))A \\
\end{align*}

\subsubsection{Default \textit{augmented} Dataset} 

By default in all of our experiments we add synthetically generated proofs to the dataset extracted from \textit{set.mm} as shown in Table~\ref{table:augmented}. We'll denote this dataset as our \textit{augmented} dataset. The synthetically generated proofs account for approximately 1\% of our training data which empirically appeared as big enough to achieve decent performance on the tasks we cared about and small enough not to hurt performance on the \textit{valid} set, especially for small models. We attempted scaling the portion of synthetic proofs to 5\% of the dataset and found out that it hurt performance for the model sizes we studied. It is nonetheless possible that including more synthetic data may turn out to be beneficial for larger models than the ones studied in this paper.

\begin{table}[ht]
\caption{Number of proofs and proofsteps adjunct to constitute our \textit{augmented} dataset.}
\centering
\begin{tabular}{ |l|c|c| }
    \hline
    Generator & Number of Proofs & Number of Proofsteps \\
    \hline
    9-digit Addition (in $\mathbb{Z}$) & 100 & 4541 \\
    9-digit Division & 100 & 10047 \\
    9-digit Modulo & 50 & 4438 \\
    9-digit Exponentiation & 50 & 910 \\
    Ring Equalities ($\mathit{depth}=6$, $\mathit{nbvar}=2$) & 50 & 1373 \\
    Ring Equalities ($\mathit{depth}=6$, $\mathit{nbvar}=3$) & 50 & 1499 \\
    \hline
 \end{tabular}
\label{table:augmented}
\end{table}

\subsection{Learned Value Function}
\label{section:learnedvalue}

To achieve better performance, we also iteratively train a value function to guide the proof search, in place of the cumulative logprob priority described above.

We implement the value function by means of an \textit{outcome objective} as follows. Any time we attempt to prove a statement, we will generate a significant number of intermediate goals. Some of these goals will lead to the proof, other goals will be proved without being part of the final proof, while others will not be resolved at all. To obtain a value function, we simply train our model to predict whether a goal produced during proof search ended up being resolved by generating a new dataset of the following form:  

\begin{verbatim}
GOAL <GOAL> OUTCOME <P|N><EOT>
\end{verbatim}

Where a goal ends with a "P" if was resolved, and "N" otherwise.

The binary nature of the \textit{OUTCOME} allows the definition of a \textit{provability function} $f_P$ as the conditional probability of token \verb|P| given a \textit{GOAL} without having to introduce a separate value head. Given a goal $g$, for a model parametrized by $\theta$:

$$
f_{P}^{\theta}(g) = p_{\theta}(\textit{"P"}|g) \approx_{\textit{trained}}
1-p_{\theta}(\textit{"N"}|g)
$$

We then define our value function $V$ on goals with:

$$
V^{\theta}(g) = \prod_{g' \in \textit{siblings(g)}} f_{P}^{\theta}(g')
$$

Not having to introduce a separate value head greatly simplifies the overall architecture. Training only involves augmenting the dataset with \textit{outcome objectives} (as additional masked sentences) and sampling the "provability" function simply consists in reading the probability distribution for the token following the \verb|OUTCOME| keyword (which can be done in one forward pass).

\subsubsection{Iterative Data Generation and Training}

Having access to a formal verifier enables us to generate the training data for $f_P$ in a fully synthetic manner by first training a model on the \textit{proofstep objective}, then sampling proofs (using cumulative logprob priority) for statements from the training set, and finally, annotating goals visited by the proof searches as positives if they were proved and as negatives otherwise.

These annotations are used to train $f_P$ and the entire process can be run iteratively, similarly to \textit{Expert Iteration}~\cite{anthony2017thinking}, sampling proofs using the newly trained $V$ (instead of cumulative logprob) to guide proof search for subsequent iterations.

At each iteration we entirely re-train the model on both objectives at the same time on the dataset constructed as follows:

\begin{itemize}
    \item  We extract the full proofs that were found for statements from the training set at each previous iteration and merge them with the original training set. We deduplicate proof steps at the proof level. This dataset becomes our new train set for the \textit{proofstep objective}.
    \item We extract the annotated goals visited by the proof searches for statements from the train set as well as the goals from the original train set (annotated positively) and deduplicate the goals giving priority to positive outcomes annotations. This dataset becomes our new train set for the \textit{outcome objective}.
\end{itemize}

This iterative training allows controlling for overfitting on both objectives by processing in the same way the data generated by proof searches on statements from the \textit{valid} set and using the resulting datasets to track their associated \textit{valid} loss.

Training a value function gives an opportunity to the model to learn from its errors on data it generates. It also shifts proof searches from breadth first exploration to one that is more focused, adaptively based on the level of confidence modeled by $V$.

\section{Experiments}
\label{section:experiments}

We fine-tune all of our models with 1024 examples per global batch and a context size of 2048 tokens, for at most 32B tokens (our \textit{augmented} dataset contains $\sim 1B$ tokens), early stopping at min valid loss when applicable. We anneal the learning-rate to zero (over 32B tokens). We found that restarting the training with an annealing to zero that matches the early-stopping for a given model only provides a marginal improvement, and avoided doing so.

The models are trained with the BPE encoding reported in \cite{brown2020language}, the same tokenization being used for text, code or formalized statements. We leave as future work a thorough ablation of the encoding as preliminary experimental results demonstrate possible gains with specialized tokenization techniques. 

\subsection{Baselines}

We report three baselines: (i) the state of the art for Metamath's \textit{set.mm} as reported in \textit{MetaGen-IL}\cite{wang2020learning} (their methodology for benchmarking their solution is close to ours so the numbers are directly comparable); (ii) a 160m parameters trained from scratch on our \textit{raw} dataset using the \textit{proofstep objective}; and (iii) a 160m parameters trained from scratch on our \textit{augmented} dataset (same objective).

\begin{table}[ht]
\caption{Baseline performance from \textit{MetaGen-IL} as well as a 160m parameters model trained on the \textit{raw} and \textit{augmented} datasets.} 
\centering
\begin{tabular}{ |l|c|c| }
    \hline
    Model & Performance & \# Tokens \\
    \hline
    \textit{MetaGen-IL}\cite{wang2020learning} & 21.16\% & N/A \\
    160m \textit{raw dataset} (\textit{ours}) & \textbf{29.22\%} & 18B \\
    160m \textit{augmented dataset} (\textit{ours}) & 28.96\% & 18B \\
    \hline
\end{tabular}
\label{table:baselines}
\end{table}

We explain the improvement over \textit{MetaGen-IL} (despite not relying on forward proving data generation techniques) by our use of a simpler architecture (one unique Transformer vs 3 separate GRU networks); a more straightforward objective (direct auto-regressive generation of the full tactic as text vs separate premise selection and generation of the substitutions); more learnable parameters (160m vs ~300k (3 2-layers bi-directional GRUs with 128 hiddens)); and more compute at training as well as test time.

Note that the dataset augmentation may have a marginal negative effect on performance on the \textit{valid} set with our 160m model (but we're within typical variance). We report in section~\ref{section:samplecomplexity} a more reliably positive effect with a pre-trained 700m model.

\subsection{Model Size}

\begin{table}[ht]
\caption{Performance for various model sizes trained on the \textit{augmented} datasets.} 
\centering
\begin{tabular}{ |l|c|c|c| }
    \hline
    Model & Performance & Perplexity & \# Tokens \\
    \hline
    160m \textit{augmented} & 28.96\% & 1.041 & 18B \\
    400m \textit{augmented} & 30.23\% & 1.042 & 18B \\
    700m \textit{augmented} & \textbf{31.58\%} & \textbf{1.040} & 18B \\
    \hline
\end{tabular}
\label{table:modelsize}
\end{table}
    
These results demonstrate that model size positively impacts performance in our formal setup, despite the training dataset being limited in size (we train for $\sim 18$ epochs). Note that the bigger the model the more compute we use at training time as well as benchmarking.

\subsection{Pre-training}

Models are pre-trained on CommonCrawl using GPT-3's\cite{brown2020language} methodology for 260B tokens. When studying the effect of pre-training on \textit{WebMath} we start from a \textit{CommonCrawl} pre-trained model and continue pre-training on \textit{WebMath} for 16B additional tokens. We also report results after pre-training on \textit{GitHub} only instead of \textit{WebMath} for the same number of tokens.  

\begin{table}[ht]
\caption{Performance for various model sizes and pre-training datasets.} 
\centering
\begin{tabular}{ |l|c|c|c| }
    \hline
    Model & Performance & Perplexity & \# Tokens \\
    \hline
    160m \textit{from scratch} & 28.96\% & 1.041 & 18B \\
    160m \textit{CommonCrawl} & 32.34\% &  1.030 & 16B \\
    160m \textit{Github} & 33.61\% & 1.030 & 16B \\
    160m \textit{WebMath} & 34.79\% & 1.029 & 16B \\
    700m \textit{from scratch} & 31.58\% & 1.040 & 18B \\
    700m \textit{CommonCrawl} & 39.61\% & 1.026 & 15B \\
    700m \textit{Github} & 41.55\% & 1.025 & 15B \\
    700m \textit{WebMath} & \textbf{42.56\%} & \textbf{1.024} & 15B \\
    \hline
\end{tabular}
\label{table:pretrain}
\end{table}

We hypothesize that the positive pre-training effect is primarily driven by the emergence and transfer of features that are relevant to formal reasoning. It is possible to argue that most of these features are probably shallow and mostly relevant at the syntactical level but the lower performance achieved with Github only in comparison to WebMath suggests that some features may be more elaborate. We leave as future work a broader investigation of this question, which could be achieved by studying the performance of linear probes on the
features of the different pre-trained models with respect to a formal objective, such as the truthiness of a set of statements provided in the Metamath (or any other formal) language.

\begin{table}[ht]
\caption{Performance for model sizes ranging from 160m to 1.5b parameters, pre-trained on \textit{WebMath}.} 
\centering
\begin{tabular}{ |l|c|c|c| }
    \hline
    Model & Performance & Perplexity & \# Tokens \\
    \hline
    160m (\textit{WebMath}) & 34.79\% & 1.029 & 16B \\
    400m (\textit{WebMath}) & 39.94\% & 1.026 & 15B \\
    700m (\textit{WebMath}) & \textbf{42.56\%} & 1.024 & 15B \\
    1p5b (\textit{WebMath}) & 42.39\% & 1.024 & 13B \\
    \hline
\end{tabular}
\label{table:webmath}
\end{table}

It is unclear why we do not observe a smooth improvement in performance between the 700m and the 1.5b models in table~\ref{table:webmath}. The lack of guarantee that the \textit{valid} set has a smooth difficulty pattern may play a role here. Another effect may originate from the limited size of the training set, leading the training dynamics to saturate as we grow the number of parameters. We leave as future work a closer study of this effect which could be accomplished by training on various fractions of the training dataset and checking for similar saturation plateaux.

\subsection{Learned Value Function}

We report the performance of our models as we iteratively train on data generated by sampling proofs against the verifier.

\begin{table}[ht]
\caption{Performance of the 160m and 700m parameters models as we iterate through the learned value function data generation and re-training process. \textit{policy only} consists in adding new positive proofs found to the policy training (without training a value function) while \textit{policy+value} consists in the full iterative data-generation and training described in section~\ref{section:learnedvalue}.} 
\centering
\begin{tabular}{ |l|c|c|c| }
    \hline
    Model & Iteration 0 & Iteration 1 & Iteration 2 \\
    \hline
    160m (\textit{WebMath}) \textit{policy only} & 34.79\% & 38.17\% & 38.34\% \\
    160m (\textit{WebMath}) \textit{policy+value} & & 39.27\% & 40.70\% \\
    700m (\textit{WebMath}) \textit{policy only} & 42.56\% & 42.23\% & 43.15\% \\
    700m (\textit{WebMath}) \textit{policy+value} & & 44.59\% & \textbf{47.21\%} \\
    \hline
\end{tabular}
\label{table:value}
\end{table}

While overfitting on the train set does not generally appear to negatively impact performance on the \textit{valid} set (and can even often help noticeably if not too catastrophic), we discovered that it dramatically hurts our iterative training process. We hypothesize that overfitting collapses the data generation in a mode where exploration is weakened, the model being overly optimistic about its predictions on the \textit{train} set. We therefore carefully avoid overfitting by tracking the loss on the associated \textit{valid} set, early stopping as we reach a minimum.

There is probably additional performance to be extracted by running more iterations given how continuous this iterative process appears to be. We leave as future work the design of an iterative data generation process that is less compute intensive. Indeed, we believe that a lot of computation is spent on subgoals that are not necessarily providing a lot of signal for the value function, and each iteration is quite compute intensive as it requires sampling proofs for the entire training set (which takes $\sim 20k$ GPU.hours on V100s in our current setup).

\subsection{Sample Complexity}
\label{section:samplecomplexity}

Ablation of our synthetic dataset augmentation demonstrates that synthetically generated proofs generalize to some extent and provide a noticeable uplift in performance on the \textit{valid} set for larger models.

\begin{table}[ht]
\caption{Ablation of the \textit{augmented} dataset for 160m and 700m parameters models.} 
\centering
\begin{tabular}{ |l|c|c|c| }
    \hline
    Model & Performance & Perplexity & \# Tokens \\
    \hline
    160m \textit{(WebMath)} \textit{raw dataset} & 34.12\% & 1.029 & 16B \\
    160m \textit{(WebMath)} \textit{augmented dataset} & 34.79\% & 1.029 & 16B \\
    700m \textit{(WebMath)} \textit{raw dataset} & 40.28\% & 1.024 & 15B \\
    700m \textit{(WebMath)} \textit{augmented dataset} & 42.56\% & 1.024 & 15B \\
    \hline
\end{tabular}
\label{table:augmentedabl}
\end{table}

Our main motivation for including synthetic proofs in our training, beyond the relative uplift achieved, is the study of the effect of model size and training a value function on the sample complexity of our models, as we control exactly how many examples from the synthetic domain we use for training. Table~\ref{table:samplecomplexity} reports the performance on 100 synthetically generated statements (different from the train set) as well as the number of synthetic proofs present in the training set for each model (in parenthesis).

\begin{table}[ht]
\caption{Performance of our models on 100 test statements from our synthetic generators (run with the same parameters used to augment the training set (see table~\ref{table:augmented}).} 
\centering
\begin{tabular}{ |l|c|c|c| }
    \hline
    Model & 9-digit addition & 9-digit division & Ring equalities \\
    \hline
    160m \textit{raw} & 13\% (0) & 4\% (0) & 6\% (0) \\
    160m \textit{augmented} & 78\% (100) & 27\% (100) & 77\% (100) \\
    160m \textit{policy+value} (iteration 1) & 87\% (100) & 24\% (100) & 71\% (100) \\
    160m \textit{policy+value} (iteration 2) & 90\% (100) & 28\% (100) & 79\% (100) \\
    700m \textit{raw} & 12\% (0) & 5\% (0) & 7\% (0) \\
    700m \textit{augmented} & 76\% (100) & 32\% (100) & 82\% (100) \\
    700m \textit{policy+value} (iteration 1) & 90\% (100) & 40\% (100) & 78\% (100) \\
    700m \textit{policy+value} (iteration 2) & \textbf{92\% (100)} & \textbf{47\% (100)} & \textbf{88\% (100)} \\
    \hline
\end{tabular}
\label{table:samplecomplexity}
\end{table}

This demonstrates the close (yet not perfectly correlated) relationship between sample complexity and performance in our formal reasoning setup, suggesting that sample complexity is an important driver of improved performance with formal mathematics.

More importantly it demonstrates that our models are capable of acquiring new non-trivial capabilities with a number of training examples that is compatible with manual formalization. We plan in the future to study similar learning dynamics for more challenging tasks for which we don't have a synthetic generator.

\subsection{Results}

\begin{table}[ht]
\caption{Performance of our 700m model \textit{policy+value} (iteration 2) as we double the number of attempts $a$ per proposition (with $d=256$).} 
\centering
\begin{tabular}{ |l|c|c| }
    \hline
    Attempts & Performance & Delta \\
    \hline
    $a=2$ & 42.90\% & \\
    $a=4$ & 47.29\% & +4.39\% \\
    $a=8$ & 51.26\% & +3.97\% \\
    $a=16$ & 54.05\% & +2.99\% \\
    $a=32$ & 56.50\% & +2.45\% \\
    \hline
\end{tabular}
\label{table:attempts}
\end{table}

We attempted to push the performance of our models by increasing both the number of expansions per proof search from $d=128$ to $d=256$, and the number of attempts per proofs from $a=4$ to $a=32$. We report the achieved performance as a function of the number of attempts per statements on the \textit{valid} set in Table~\ref{table:attempts}.

Finally, we performed a final evaluation with $d=256$ and $a=32$ of our 700m model \textit{policy+value} (iteration 2) on the held-out \textit{test} set:

$$
\text{Perf}_{a=32,e=32,d=256}^{\mathit{test}}(\theta_{\textit{700m}}) = 56.22\%
$$

\section{Output}

We describe in this section two projects we executed, aimed at sharing with the Metamath community results and tools based on our work.

\subsection{Proof Shortening}

We contributed 23 shortened proofs\footnote{\url{https://github.com/metamath/set.mm/pull/1547}}\footnote{\url{https://github.com/metamath/set.mm/pull/1561}} of theorems to the Metamath library. These proofs were generated by the \textit{GPT-f} automated prover. To discover shorter proofs, we sampled proofs for statements from the \textit{set.mm} library, comparing the length of the solutions found by our models to their ground truth versions, also verifying that the shorter proofs didn't rely on additional axioms.

The reception\footnote{\url{https://groups.google.com/g/metamath/c/-FNsw2wyllI}} from the Metamath community was positive, proof length being a metric the community care about:

\begin{displayquote}
“I had a look at the proofs—very impressive results! Especially because we had a global minimization recently, and your method found much shorter proofs nevertheless.”
\end{displayquote}

\begin{displayquote}
“Any ML-based system is impressive if it can find many shorter proofs than the ones we already have. Nice work.”
\end{displayquote}

\begin{displayquote}
“The shorter proof is easier to translate. It’s more symmetric in that it treats A and B identically. It’s philosophically more concise in that it doesn’t rely on the existence of a universal class of all sets.”
\end{displayquote}

To our knowledge, these shortened proofs are the first effective contribution of a deep learning system to a formal mathematics library\footnote{To determine whether other deep learning-based provers have made contributions to their respective libraries, we looked for such contributions in the following systems: Holist family in HOL Light, CoqGym+ASTatic in Coq, TacticToe in HOL4. In addition, we interviewed 6 experts in formal mathematics and/or deep learning applied to formal mathematics.}

\subsection{\textit{GPT-f} Proof Assistant}
\label{section:gptfproofassistant}

We created an online proof assistant\footnote{\url{https://groups.google.com/g/metamath/c/D09W2QVR-_I/m/g_rsqGj0AAAJ}} to allow interactive proof constructions with the assistance of our models.

\begin{figure}
    \includegraphics[width=\textwidth]{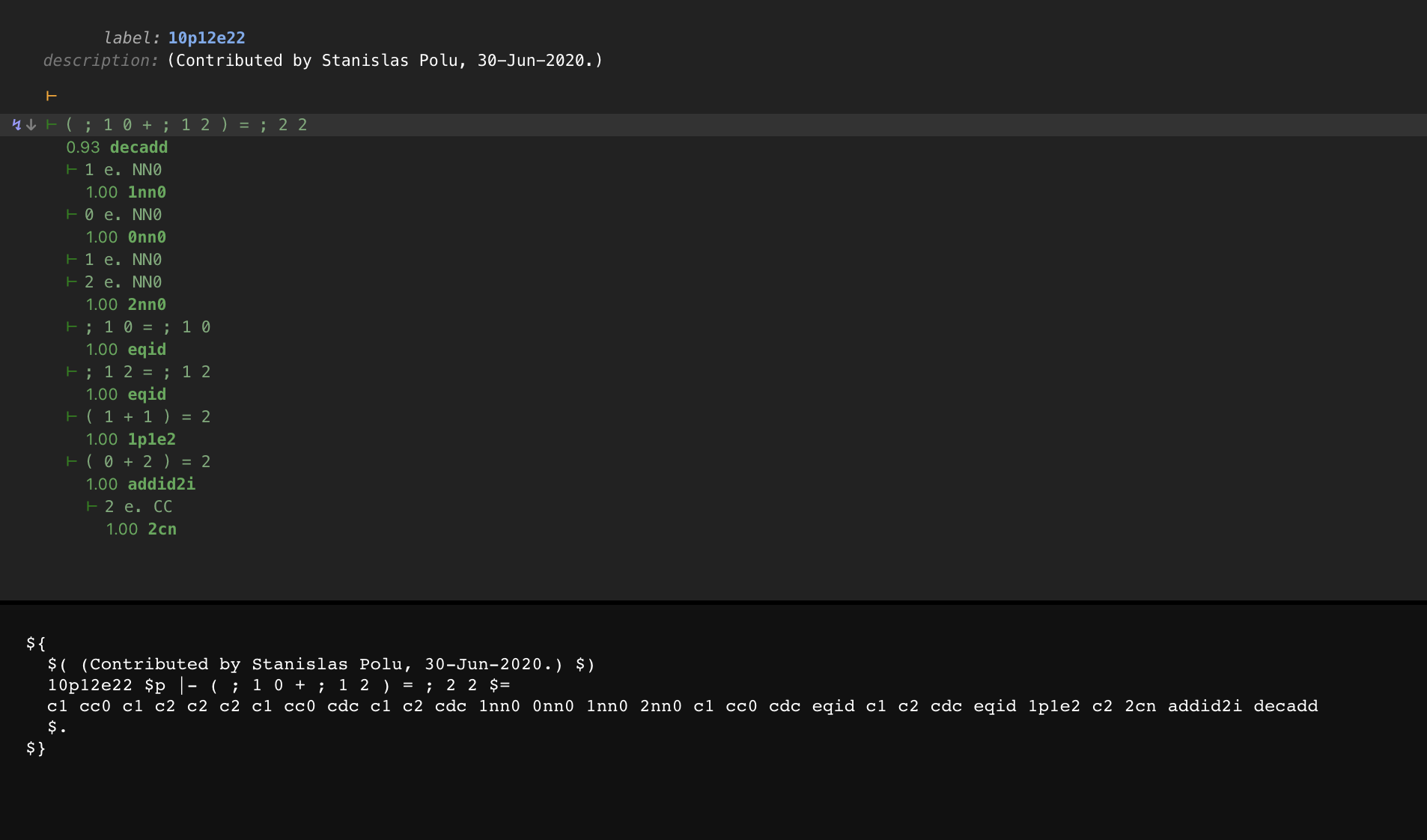}
    \caption{Screenshot of the \textit{GPT-f} Proof Assistant}
    \label{fig:mm}
\end{figure}

We used it to formalize more than 200 theorems and exercises. We found our models to be particularly useful to automatically generate a variety of technical low level proofsteps required in most Metamath proofs, search the library by adapting existing theorems to the format needed by the user (e.g., deduction form\footnote{\textit{Deduction Form and Natural Deduction} \url{http://us.metamath.org/mpeuni/mmnatded.html}}) and suggest theorems to use. Even when mistaken, our models generally go for the right theorems, whose erroneous substitutions are often easy to fix by humans.

We shared the proof assistant with the Metamath community with the objective for it to be mutually beneficial, helping the community to be more productive and reciprocally helping us improve our models' accuracy by automatically gathering human feedback. We also plan to extend \textit{GPT-f} to other formal systems.

\section{Conclusion}

In this paper, we present the \textit{GPT-f} automated prover and proof assistant and show that the Transformer is suitable to formal reasoning, achieving a new state of the art result on the Metamath library. In particular we demonstrate the importance of pre-training as well as iterative training of a value function. Our results suggest that tightly coupling a deep learning system with a formal system opens up interesting opportunities for further research, with the goal of better leveraging the generative power of the former and the verification capabilities of the latter.

\begin{ack}
Szymon Sidor, Jakub Pachocki, Harri Edwards, Yura Burda and Vedant Misra inspired many of the ideas presented in this work, offering their guidance throughout the process of building \textit{GPT-f}. Auguste Poiroux implemented the synthetic dataset generators presented in this paper, and formalized a large number of theorems using the proof assistant, providing invaluable feedback in the process. Szymon Sidor, Pranav Shyam, John Schulman, Jared Kaplan, Ryan Lowe and Jack Clark slogged through drafts of this paper, identifying errors and sources of confusion as well as providing helpful suggestions. Finally, the authors would like to thank the whole Metamath community for their support, feedback, and encouragement, in particular, David A. Wheeler for his motivating enthusiasm and Mario Carneiro for his precious help on a wide variety of technical questions.
\end{ack}

\bibliographystyle{unsrt}
\bibliography{references.bib}

\appendix
\section{Key Results}

\begin{table}[ht]
\caption{Key results described in this paper (on the \textit{valid} set) with a summary of the source of performance gains.}
\centering
\begin{tabular}{ |l|c|c|l| }
    \hline
    Model & Performance & Gain & Main ablation \\
    \hline
    \textit{MetaGen-IL}~\cite{wang2020learning} & 21.16\% & & Baseline and state of the art. \\
    160m (ours) & 28.96\% & +7.8\% & Use of Transformers. \\
    700m (ours) & 31.58\% & +2.5\% & Increase in parameters count. \\
    700m \textit{WebMath} (ours) & 42.56\% & +10.9\% & Pre-training. \\
    700m \textit{policy+value} (ours) & 47.21\% & +4.6\% & Iterated learned value function. \\
    700m \textit{policy+value} $a=32$ (ours) & 56.50\% & +9.2\% & Increased test-time compute. \\
    \hline
\end{tabular}
\label{table:keyresults}
\end{table}

\section{Example Proofs Generated}

In this appendix, we display a selection of proofs generated by \textit{GPT-f} (from our \textit{valid} set). The right column contains the current goal.  The left column displays the name of the theorem applied by to the goal. Proofs are read bottom-up and the statement being demonstrated is the last goal of the table. The subgoals generated by a proof step can be retrieved by looking at the theorem names that are indented by one additional space. The statement of the theorems can be retrieved with the Metamath Proof Explorer. Substitutions are omitted for clarity, but can be inferred by looking at the statement of the theorem being applied and comparing it with the current goal and associated subgoals.

\subsection{Proof of \textbf{nn0onn0ex}}

This proof demonstrates that $n \in \mathbb{N} \land \frac{n+1}{2} \in \mathbb{N} \implies \exists m \in \mathbb{N}: n = 2m+1$. It is interesting for its first proof step. \verb|syl2anc|\footnote{\textit{Metamath Proof Explorer - syl2anc} \url{http://us.metamath.org/mpeuni/syl2anc.html}} states that assuming $P \implies Q, P \implies R, Q \land R \implies S$ then $P \implies S$. $P$ is mechanically unified with $n \in \mathbb{N} \land \frac{n+1}{2} \in \mathbb{N}$, and $S$ with $\exists m \in \mathbb{N}: n = 2m+1$ but the model freely generates substitutions for $Q$ and $R$. Looking at the subgoals, $Q$ is substituted with $\frac{n-1}{2} \in \mathbb{N}$ and $R$ with $n = 2\frac{n-1}{2}+1$ which materialises a witness for the existence of $m$.

The model is left to demonstrate $n \in \mathbb{N} \land \frac{n+1}{2} \in \mathbb{N} \implies \frac{n-1}{2} \in \mathbb{N}$, then $n \in \mathbb{N} \land \frac{n+1}{2} \in \mathbb{N} \implies n = 2\frac{n-1}{2}+1$ and finally $\frac{n-1}{2} \in \mathbb{N} \land n = 2\frac{n-1}{2}+1 \implies \exists m \in \mathbb{N}: n = 2m+1$ using the existential specialization provided by \verb|rspcev|\footnote{\textit{Metamath Proof Explorer - rspcev} \url{http://us.metamath.org/mpeuni/rspcev.html}}.

\begin{tabular}{ |l|p{105mm}| }
\hline
\verb| + nn0o       | & \verb!|- ( ( N e. NN0 /\ ( ( N + 1 ) / 2 ) e. NN0 ) -> !\newline\verb!   ( ( N - 1 ) / 2 ) e. NN0 )! \\
\verb|      + nn0cn | & \verb!|- ( N e. NN0 -> N e. CC )! \\
\verb|      + ax1cn | & \verb!|- 1 e. CC! \\
\verb|      + subcl | & \verb!|- ( ( N e. CC /\ 1 e. CC ) -> ( N - 1 ) e. CC )! \\
\verb|     + sylancl| & \verb!|- ( N e. NN0 -> ( N - 1 ) e. CC )! \\
\verb|     + 2cnd   | & \verb!|- ( N e. NN0 -> 2 e. CC )! \\
\verb|      + 2ne0  | & \verb!|- 2 =/= 0! \\
\verb|     + a1i    | & \verb!|- ( N e. NN0 -> 2 =/= 0 )! \\
\verb|    + divcan2d| & \verb!|- ( N e. NN0 -> ( 2 x. ( ( N - 1 ) / 2 ) ) = ( N - 1 ) )! \\
\verb|   + adantr   | & \verb!|- ( ( N e. NN0 /\ ( ( N + 1 ) / 2 ) e. NN0 ) -> !\newline\verb!   ( 2 x. ( ( N - 1 ) / 2 ) ) = ( N - 1 ) )! \\
\verb|  + oveq1d    | & \verb!|- ( ( N e. NN0 /\ ( ( N + 1 ) / 2 ) e. NN0 ) -> !\newline\verb!   ( ( 2 x. ( ( N - 1 ) / 2 ) ) + 1 ) = !\newline\verb!   ( ( N - 1 ) + 1 ) )! \\
\verb|    + nn0cn   | & \verb!|- ( N e. NN0 -> N e. CC )! \\
\verb|    + ax1cn   | & \verb!|- 1 e. CC! \\
\verb|    + npcan   | & \verb!|- ( ( N e. CC /\ 1 e. CC ) -> ( ( N - 1 ) + 1 ) = N )! \\
\verb|   + sylancl  | & \verb!|- ( N e. NN0 -> ( ( N - 1 ) + 1 ) = N )! \\
\verb|  + adantr    | & \verb!|- ( ( N e. NN0 /\ ( ( N + 1 ) / 2 ) e. NN0 ) -> !\newline\verb!   ( ( N - 1 ) + 1 ) = N )! \\
\verb| + eqtr2d     | & \verb!|- ( ( N e. NN0 /\ ( ( N + 1 ) / 2 ) e. NN0 ) -> !\newline\verb!   N = ( ( 2 x. ( ( N - 1 ) / 2 ) ) + 1 ) )! \\
\verb|    + oveq2   | & \verb!|- ( m = ( ( N - 1 ) / 2 ) -> ( 2 x. m ) = !\newline\verb!   ( 2 x. ( ( N - 1 ) / 2 ) ) )! \\
\verb|   + oveq1d   | & \verb!|- ( m = ( ( N - 1 ) / 2 ) -> ( ( 2 x. m ) + 1 ) = !\newline\verb!   ( ( 2 x. ( ( N - 1 ) / 2 ) ) + 1 ) )! \\
\verb|  + eqeq2d    | & \verb!|- ( m = ( ( N - 1 ) / 2 ) -> ( N = ( ( 2 x. m ) + 1 ) !\newline\verb!   <-> N = ( ( 2 x. ( ( N - 1 ) / 2 ) ) + 1 ) ) )! \\
\verb| + rspcev     | & \verb!|- ( ( ( ( N - 1 ) / 2 ) e. NN0 /\ !\newline\verb!   N = ( ( 2 x. ( ( N - 1 ) / 2 ) ) + 1 ) ) -> !\newline\verb!   E. m e. NN0 N = ( ( 2 x. m ) + 1 ) )! \\
\verb|+ syl2anc     | & \verb!|- ( ( N e. NN0 /\ ( ( N + 1 ) / 2 ) e. NN0 ) -> !\newline\verb!   E. m e. NN0 N = ( ( 2 x. m ) + 1 ) )! \\
\hline
\end{tabular}

Such generation of exogenous terms, here to demonstrate an existence proof, is exactly what motivated our work. It's therefore encouraging to witness it effectively happening in practice.

\subsection{Proof of \textbf{uznn0sub}}

This proof demonstrates that $n \ge m \in \mathbb{Z} \implies (n-m) \in \mathbb{N}$. It exhibits another form of term generation. Here, \verb|sylibr|\footnote{\textit{Metamath Proof Explorer - sylibr} \url{http://us.metamath.org/mpeuni/sylibr.html}} states that assuming $P \implies Q, R \Leftrightarrow Q$ then $P \implies R$. Again, $P$ is mechanically unified to $n \ge m \in \mathbb{Z}$, and $R$ with $(n-m) \in \mathbb{N}$. The model is left to generate freely a substitution for $Q$: $(n-m) \in \mathbb{Z} \land 0 \le (n-m)$. The equivalence $R \Leftrightarrow Q$ to demonstrate becomes $(n-m) \in \mathbb{N} \Leftrightarrow (n-m) \in \mathbb{Z} \land 0 \le (n-m)$ which is exactly the statement of a theorem available in the Metamath library, \verb|elnn0z|\footnote{\textit{Metamath Proof Explorer - elnn0z} \url{http://us.metamath.org/mpeuni/elnn0z.html}}. The statement of \verb|elnn0z| is memoized by the model, and the generation of the substitution term for $Q$ is driven by this memoization.

\begin{tabular}{ |l|p{105mm}| }
\hline
\verb!   + eluzelz  ! & \verb!|- ( N e. ( ZZ>= ` M ) -> N e. ZZ )! \\
\verb!   + eluzel2  ! & \verb!|- ( N e. ( ZZ>= ` M ) -> M e. ZZ )! \\
\verb!  + zsubcld   ! & \verb!|- ( N e. ( ZZ>= ` M ) -> ( N - M ) e. ZZ )! \\
\verb!   + eluzle   ! & \verb!|- ( N e. ( ZZ>= ` M ) -> M <_ N )! \\
\verb!    + eluzelre! & \verb!|- ( N e. ( ZZ>= ` M ) -> N e. RR )! \\
\verb!     + eluzel2! & \verb!|- ( N e. ( ZZ>= ` M ) -> M e. ZZ )! \\
\verb!    + zred    ! & \verb!|- ( N e. ( ZZ>= ` M ) -> M e. RR )! \\
\verb!   + subge0d  ! & \verb!|- ( N e. ( ZZ>= ` M ) -> ( 0 <_ ( N - M ) <-> M <_ N ) )! \\
\verb!  + mpbird    ! & \verb!|- ( N e. ( ZZ>= ` M ) -> 0 <_ ( N - M ) )! \\
\verb! + jca        ! & \verb!|- ( N e. ( ZZ>= ` M ) -> ( ( N - M ) e. ZZ /\ !\newline\verb!   0 <_ ( N - M ) ) )! \\
\verb! + elnn0z     ! & \verb!|- ( ( N - M ) e. NN0 <-> ( ( N - M ) e. ZZ /\ !\newline\verb!   0 <_ ( N - M ) ) )! \\
\verb!+ sylibr      ! & \verb!|- ( N e. ( ZZ>= ` M ) -> ( N - M ) e. NN0 )! \\
\hline
\end{tabular}

\subsection{Proof of \textbf{pm4.78}}

This proof displays the model capabilities to demonstrate non-trivial propositional logic statements, a task of interest because of its relationship to SAT solving.

\begin{tabular}{ |l|p{105mm}| }
\hline
\verb!    + pm2.21! & \verb!|- ( -. ph -> ( ph -> ps ) )! \\
\verb!   + orcd   ! & \verb!|- ( -. ph -> ( ( ph -> ps ) \/ ( ph -> ch ) ) )! \\
\verb!    + ax-1. ! & \verb!|- ( ps -> ( ph -> ps ) )! \\
\verb!    + ax-1  ! & \verb!|- ( ch -> ( ph -> ch ) )! \\
\verb!   + orim12i! & \verb!|- ( ( ps \/ ch ) -> ( ( ph -> ps ) \/ ( ph -> ch ) ) )! \\
\verb!  + ja      ! & \verb!|- ( ( ph -> ( ps \/ ch ) ) -> ( ( ph -> ps ) \/ !\newline\verb!   ( ph -> ch ) ) )! \\
\verb!    + orc   ! & \verb!|- ( ps -> ( ps \/ ch ) )! \\
\verb!   + imim2i ! & \verb!|- ( ( ph -> ps ) -> ( ph -> ( ps \/ ch ) ) )! \\
\verb!    + olc   ! & \verb!|- ( ch -> ( ps \/ ch ) )! \\
\verb!   + imim2i ! & \verb!|- ( ( ph -> ch ) -> ( ph -> ( ps \/ ch ) ) )! \\
\verb!  + jaoi    ! & \verb!|- ( ( ( ph -> ps ) \/ ( ph -> ch ) ) -> !\newline\verb!   ( ph -> ( ps \/ ch ) ) )! \\
\verb! + impbii   ! & \verb!|- ( ( ph -> ( ps \/ ch ) ) <-> ( ( ph -> ps ) \/ !\newline\verb!   ( ph -> ch ) ) )! \\
\verb!+ bicomi    ! & \verb!|- ( ( ( ph -> ps ) \/ ( ph -> ch ) ) <-> !\newline\verb!   ( ph -> ( ps \/ ch ) ) )! \\
\hline
\end{tabular}

\end{document}